\documentclass{article}
\usepackage{ijcai16}

\usepackage{times}
\usepackage[linesnumbered,ruled]{algorithm2e}
\usepackage{pifont}
\usepackage{caption}
\usepackage{subcaption}
\usepackage{amsmath}
\usepackage{amsthm}
\usepackage{amsfonts}
\usepackage{commath}
\usepackage{nicefrac}
\usepackage{float}
\newcommand{\argmin}{\mathop{\mathrm{arg min}}\limits}
\newcommand{\argmax}{\mathop{\mathrm{argmax}}\limits}
\usepackage{graphicx}
\graphicspath{ {figs/} }
\newtheorem{theorem}{Theorem}[section]

\newtheorem{lemma}[theorem]{Lemma}

\title{An Online Mechanism for Ridesharing in 
 Autonomous Mobility-on-Demand Systems}
\author{Wen Shen \and Cristina V. Lopes\\ 
University of California, Irvine  \\
 Irvine, CA 92697, USA\\
\{wen.shen, lopes\}@uci.edu
\And
Jacob W. Crandall\\
Masdar Institute of Science and Technology\\
 Abu Dhabi, PO Box 54224, UAE\\
jcrandall@masdar.ac.ae
}

\begin{document}

\maketitle

\begin{abstract}
With proper management, Autonomous Mobility-on-Demand (AMoD) systems have great potential to satisfy the transport demands of urban populations by providing safe, convenient, and affordable ridesharing services.  Meanwhile, such systems can substantially decrease private car ownership and use,  and thus significantly reduce traffic congestion, energy consumption and carbon emissions.  To achieve this objective, an AMoD system requires private information about the demand from passengers.  However, due to self-interestedness, passengers are unlikely to cooperate with the service providers in this regard.  Therefore, an online mechanism is desirable if it incentivizes passengers to truthfully report their actual demand.  For the purpose of promoting ridesharing, we hereby introduce a posted-price,  integrated online ridesharing mechanism (IORS) that satisfies desirable properties such as ex-post incentive compatibility, individual rationality and budget-balance.  Numerical results indicate the competitiveness of IORS compared with two benchmarks, namely the optimal assignment and an offline, auction-based mechanism.
\end{abstract}

\section{Introduction}
\begin{figure}[h]
\centering
\includegraphics[width=.55\linewidth]{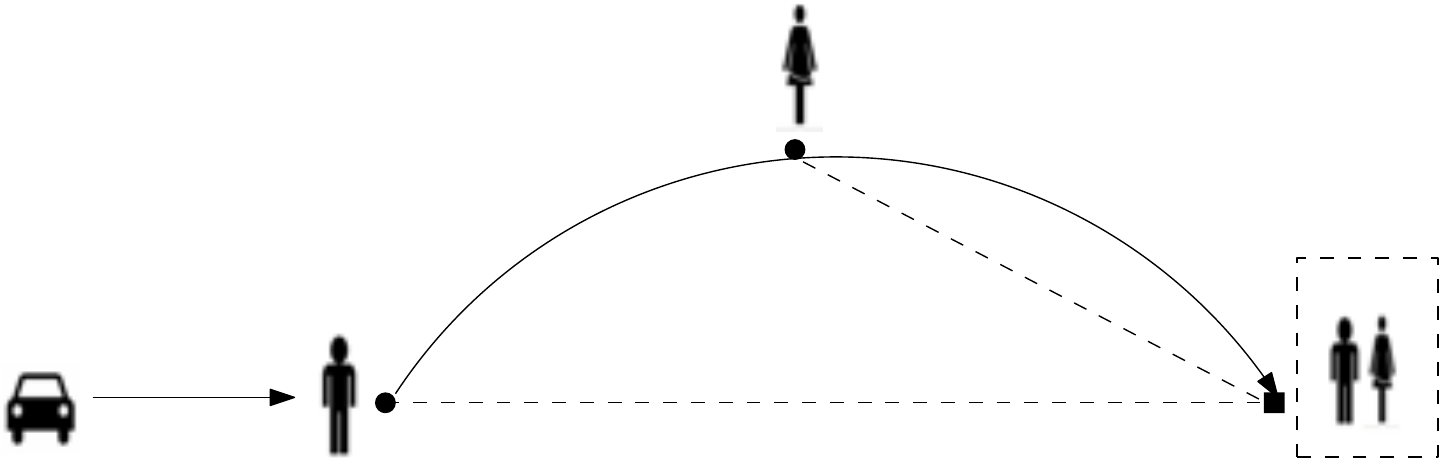}
\caption{A simple scenario of ridesharing: dashed lines indicate the real demand; solid lines with arrow indicate actual routes.}
\label{fig:ridesharingsim}
\end{figure}

The rise of private car ownership and use has brought many social and environmental challenges,  including traffic congestion,  increased greenhouse gas emissions~\cite{poudenx2008effect}.  One possible solution to address the challenges is to promote ridesharing~\cite{furuhata2013ridesharing,caulfield2009estimating,levofsky2001organized} among passengers by providing incentives (e.g., lower fares for the shared trips than individual trips) to them. In such a scenario, a limited number (depending on the seat capacity of the vehicle) of passengers who have similar itineraries share a ride and split the fares.  Ridesharing (as shown in Figure~\ref{fig:ridesharingsim}) increases the occupancy of vehicles during traveling, making it possible to transport more passengers with fewer vehicles running on the roads. 

Rideshare experiences can be significantly improved as  autonomous vehicle technologies mature. This is  because AMoD systems have the potential to provide safe, convenient, and affordable mobility solutions for passengers,  while reducing greenhouse emissions and private car ownership~\cite{mitchell2010reinventing,chong2013autonomy,shen2015managing,spieser2014toward}. Unlike traditional mobility-on-demand systems (e.g., taxis, shuttles),  an AMoD system is equipped with a fleet of self-driving, electric cars with no drivers needed. This enables seamless cooperation between the information center and the autonomous vehicles (AVs).

Several mechanisms~\cite{kleiner2011mechanism,kamar2009collaboration,cheng2014mechanisms,zhao2015incentive}  have been introduced to promote ridesharing in traditional mobility-on-demand systems. These mechanisms require passengers to directly reveal the valuation of the rides. While interesting and insightful from a theoretical perspective, they may not work well in practice since they require passengers to reveal the exact value of their rides, which could be problematic~\cite{naor1999privacy,larson2001costly,babaioff2015dynamic}. In this case, posted-price mechanisms are more appealing because passengers only need to accept offers with value greater than the posted price, without revealing their actual valuations to service providers.

Some of the mechanisms require additional constraints (e.g., dual ride shares only, linear in commitment) to satisfy desirable properties such as strategy-proofness and budget-balance~\cite{kleiner2011mechanism,zhao2015incentive}. Besides, many mechanisms assume that passengers are only motivated by monetary incentives (i.e., lower fares)~\cite{kleiner2011mechanism,kamar2009collaboration,cheng2014mechanisms,zhao2015incentive}. They neglect the fact that non-monetary factors such as time, comfortability and privacy, are also important, or even critical when people make decisions on whether to use the service or not. 

Another drawback of these mechanisms is that they process the ride requests in batch and do not work in online environment~\cite{kleiner2011mechanism,kamar2009collaboration,cheng2014mechanisms,zhao2015incentive}. In AMoD systems, service providers are committed to offering an immediate response to each request sent by passengers via a smart device. Besides, they assume that the demand is fixed without consideration of the dynamic nature of demand responsive systems.

A desirable mechanism is expected to be truthful and online~\cite{parkes2004mdp,gallien2006dynamic,nisan2007algorithmic}. It should be able to provide a fare quote immediately after the submission of a request. It needs to consider major non-monetary factors (e.g., latest departure time) as well as the dynamic nature of demand-driven systems. However, such a mechanism is yet to be designed. To bridge the gap and transcend conventional transport models like private car ownership, we introduce a truthful online mechanism called IORS for AMoD systems. We implement a simple, abstracted, yet powerful simulator that enables efficient modeling of ridesharing in AMoD systems.  Numerical results show that the IORS mechanism outperforms the cutting-edge auction-based mechanism for last-mile mobility systems~\cite{cheng2014mechanisms} substantially. It has a very close performance compared to the optimal solution, but requires a shorter time to compute and requires no future knowledge about the demand.

\section{Ridesharing in AM\MakeLowercase{o}D Systems}
\begin{figure}[h]
\centering
\includegraphics[width=.35\linewidth]{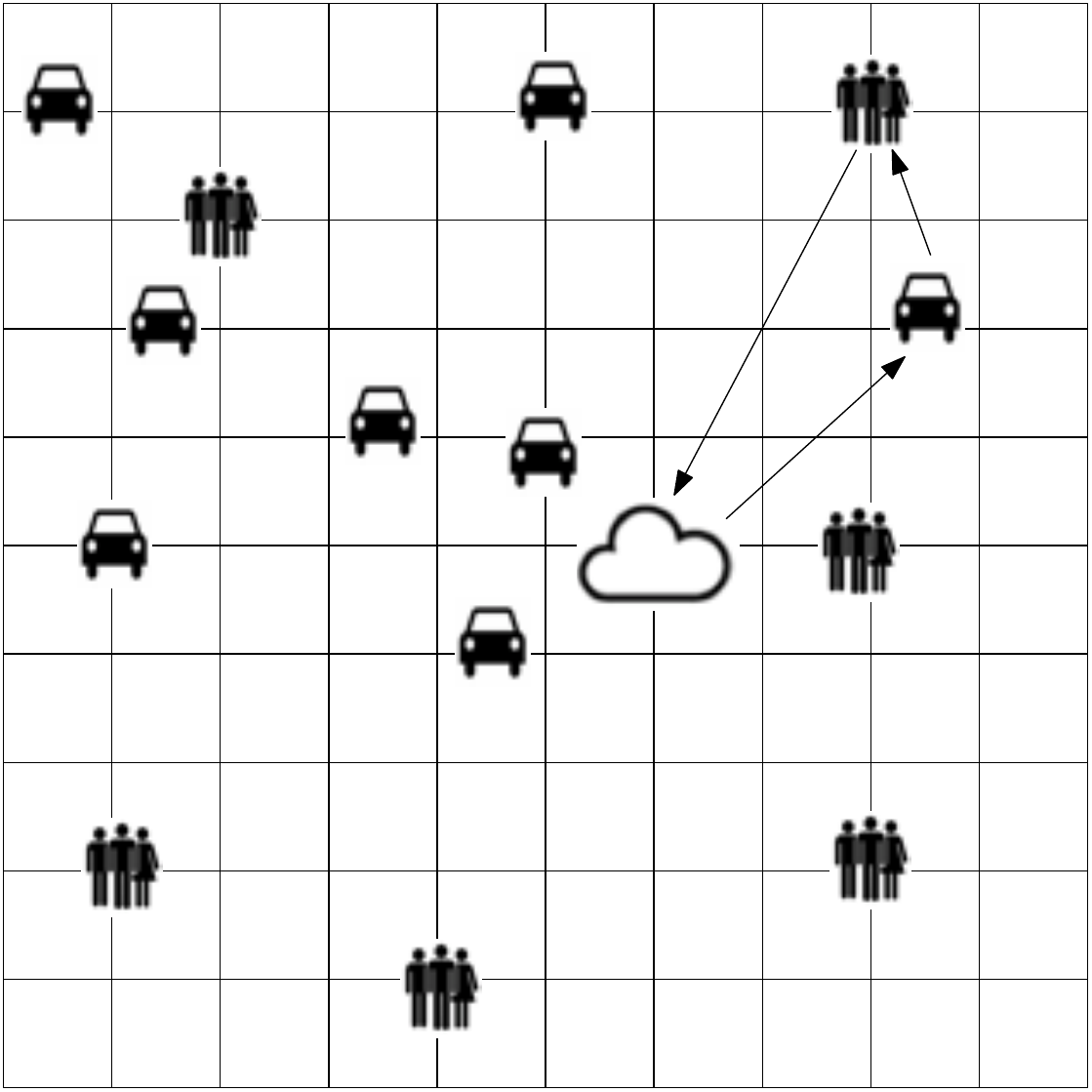}
\caption{An abstraction of ridesharing in an AMoD system operating in a grid city.}
\label{fig:abstractframework}
\end{figure}

An AMoD system (see Figure~\ref{fig:abstractframework}) can be viewed as a multi-agent system consisting of an information center, a fleet of autonomous vehicle agents,  and self-interested passengers who dynamically enter and exit the system.  The working principle of the AMoD system is straightforward:  when a passenger needs a ride, she sends the ride request to the information center using a smart device. This initiates the demand for mobility. The information center next computes a fare quote and sends it to the passenger. If the passenger accepts the fare estimate, the information center then calculates an assignment.  As long as a plan has been calculated, it will be sent to both the AV assigned and the passenger who has just submitted the request.  Both the passenger and the AV are committed to executing it.  Otherwise, the passenger will be subject to penalties. 
\section{The Online Mechanism}
\subsection{Preliminaries}
\label{sec:prelim}
In our work we consider discrete time $ \mathcal{T} = \{0, 1, 2, . . . \}$, with passengers that arrive and depart over time. Without loss of generality, we assume that the AVs never exit the AMoD system.  The information center has full knowledge of the AVs at each time.  However, passengers' demand information is private and hidden from the center. The mechanism designer should incentivize passengers to  truthfully reveal their demand for better system-wide optimization.

In our model we make a realistic assumption that the passengers are impatient~\cite{horn2002fleet}. They will leave the AMoD system and switch to other alternatives if the pickup time is later than their latest departure time. 

Let $\mathcal{I}^{t}$ denote a group of passengers who have mobility demand at time $t \in \mathcal{T}$. 
At each time, a passenger $i \in \mathcal{I}^{t}$  submits a request $r_{i}^{t} \in \mathcal{R}^{t}$ to the information center. The request $r_{i}^{t}$ can be represented as a tuple $(o_{i}, d_{i}, t_{i}, \bar{t}_{i})$, where $ o_{i}$, $ d_{i}$, $t_{i}$ and $\bar{t}_{i} $ are passenger $i$'s origin, destination, arrival time and latest departure time, respectively. Here, $t_{i} = t$. To quantify the transport demand, we introduce the effective demand $\ell_{i}$, which indicates the minimum distance from passenger $i$'s origin $o_{i}$ and destination $d_{i}$. Assuming that the effective demand is independent of request submission time, we have:
\begin{equation}
\label{eq:demandind}
\forall t \in \mathcal{T}, i \in \mathcal{I}^{t},  \quad \ell_{i}^{t+1} = \ell_{i}^{t} \; . 
\end{equation}
Equation~\ref{eq:demandind} indicates that the effective demand of passenger $i \in \mathcal{I}^{t}$ will not change if the passenger delays its request submission from time $t$ to time $t+1$.

Once a passenger $i$ has sent her request $r_{i}^{t}$, the system needs to provide a fare estimate $q_{r_{i}^{t}}$ to the passenger immediately, which enables the passenger to make a prompt decision on whether to accept the quote or not.  It is important to note that the quote is the upper bound of the fare rather than the final payment. The passenger only accepts the service if the quote is lower than the amount that she is willing to pay. If the passenger accepts the quote and if the system is able to provide the service given the time and routing constraints, she will be serviced with an assignment provided.  A final payment $p_{r_{i}^{t}}$ will be calculated upon the completion of the ride. Otherwise, the request will be rejected.

In our model we make the passengers who use the ride service split the operational cost of the vehicles. This enables the AMoD system to provide service without seeking for external subsidies. The mechanism designer's goal is to minimize the cost per unit effective demand. Since the operational cost is split by all the passengers, this objective reflects the social welfare. The mechanism needs to collect truthful information about the requests from passengers. However,  passengers are selfish and motivated to minimize their own cost for the rides. For instance, they may delay their request submissions, or claim a shorter waiting time to reduce their cost. Therefore, incentives should be provided to counter the manipulations. Let $C^{t}$ denote the total cost of the system incurred up to now from $t=0$, and $\mathcal{I^{\prime}}^{t} \subseteq \mathcal{I}^{t}$ be the set of passengers been serviced at $t$. We have the cost per demand of the system:
\begin{equation}
\mathit{W}^{\prime} = \frac{C^{t}}{\sum_{t^{\prime} \in [0, t]}\sum_{\substack{ i \in \mathcal{I^{\prime}}^{t^{\prime}}
  } }\ell_{i}} \; .
\end{equation}
Let $\mathit{W} = \nicefrac[]{1}{\mathit{W}^{\prime}}$, the mechanism designer's goal is equivalent to maximizing the social welfare $\mathit{W}$.
Initially, the total cost is zero. That is, $C^{0} = 0$.
We assume that the total cost of the system $C^{t}$ is non-decreasing. It quantifies the minimum operational cost that the system needs to transport the passengers. Since the total cost is independent of the time and the  orders of the request submissions, the following inequation holds:
\begin{equation}
\forall t \in \mathcal{T}, \quad C^{t+1} \geq C^{t}\; .
\end{equation}
If the requests are delayed from $t$ to $t+1$:
\begin{equation}
\label{eq:independentcost}
\forall t \in \mathcal{T}, \quad C^{t+1} = C^{t}\; .
\end{equation}
Let $\delta_{r_{i}^{t}}$ denote the increase on the operational cost if passenger $i $ is serviced when she submits the request $r_{i}^{t}$ at time $t$, the following equation always holds for every $t \in \mathcal{T}$:
\begin{equation}
\label{eq:deltaequal}
C^{t+1} - C^{t} = \sum_{\substack{t^{\prime} = t+1\\ i \in \mathcal{I^{\prime}}^{t^{\prime}} }} \delta_{r_{i}^{t^{\prime}}}\; .
\end{equation}
Equation~\ref{eq:deltaequal} shows that the increase on the total operational cost from time $t$ to time $t+1$  always equals to the summation of the marginal cost that is incurred by the requests being served at time $t+1$.

Let $\mathbb{V}$ denote the set of vehicles in operation and $\mathcal{V}^{t} \subseteq \mathbb{V} $ be the set of vehicles that have at least one seat available at time $t$.  Initially, we divide the vehicles into $N$ groups (or coalitions), where $N  =\abs{\mathbb{V}}$.  At each time, a request $r \in \mathcal{R}^{t}$ is added into a group $\mathcal{X}_{v}^{t} \subseteq \mathbb{X}^{t}$ according to the mechanism policy, where $\mathbb{X}^{t}$ is the set of all groups at time $t$. At time $t$, all passengers in a group share the same fare rate (cost per unit demand).  Let $\rho$ denote the cost per unit demand, we have:
\begin{equation}
\label{eq:costrateeq}
\rho_{\mathcal{X}_{v}^{t}} = \rho_{r_{i}^{t}}\; ,
\end{equation}
where $\quad ( i \in \{ i \mid r_{i}^{t} \in \mathcal{X}_{v}^{t}, t \in \mathcal{T}\}) $.

\subsection{IORS Mechanism}

The IORS mechanism consists of three parts: fare estimation, pickup assignment and payment calculation. In the fare estimation process, the mechanism calculates a quote for each request. In the pickup assignment phase, the mechanism computes an optimal plan that minimizes the cost per unit demand. Finally, the mechanism provides payments immediately after successful completion of the rides.

\subsubsection{Fare Estimation} 
As the passengers arrive and depart dynamically, the mechanism  can only rely on the known information to compute the upper bound of the fare. The fare estimation process is described as follows (as shown in Algorithm~\ref{alg:farequote}): for each request $r^{t}_{i}$ at time $t$,  the mechanism first checks if a vehicle $v$ (with vacancies) satisfies the passenger's  latest departure time. If such a vehicle is found, then the mechanism compares the cost per unit demand before and after adding the request $r^{t}_{i}$ into the coalition, respectively. If the cost per unit demand decreases, then the fare is calculated and stored in a set $Q^{\prime}$. The mechanism selects the maximum fare in the set as the quote. Otherwise, the system rejects the request. Note that the mechanism picks the highest (instead of the lowest) cost per unit demand as the upper bound of the fare estimate. This is because the mechanism needs to adjust the assignments so that the system can produce the lowest cost per unit demand in general. Besides, it is a necessary condition for individual rationality and  incentive compatibility. The algorithm takes $\mathcal{O} (n^{3})$ time in the worst case.
\begin{algorithm}[h!]
    \SetKwInOut{Input}{Input}
    \SetKwInOut{Output}{Output}
    \SetKwInOut{Initialize}{Initialize}

    \underline{function estimate} $(t, \mathcal{R}^{t})$\;
    \Input{$t$ - Time; 
    $\mathcal{R}^{t}$ - A set of requests from passengers $\mathcal{I}^{t}$ at time $t$. }
    \Output{$\mathcal{Q}_{\mathcal{R}^{t}}$-The fare estimate for requests $\mathcal{R}^{t}$.}
 $\mathcal{Q}_{\mathcal{R}^{t}} \leftarrow \emptyset$;\\
    \While { $ r_{i}^{t} \in \mathcal{R}^{t}$} 
    {
    $\mathcal{Q}^{\prime} \leftarrow \emptyset$;\\
    \While{$v \in \mathcal{V}^{t}$} 
    {
        \tcc{$t^{\prime}$ is the estimated pickup time needed to service passenger $i^{t}$ using  vehicle $v$ }
    	Compute $t^{\prime}$;\\
    	\If{$t^{\prime} \leq \bar{t}_{i}$}
    {
    Compute $\rho_{\mathcal{X}_{v}^{t-1} \cup \{ r^{t}_{i}\}}$ ;\\
       \If{$ \rho_{\mathcal{X}_{v}^{t-1} \cup \{ r^{t}_{i}\}} < \rho_{\mathcal{X}_{v}^{t-1}}$}
       {
       		$q^{\prime} \leftarrow \ell_{i}^{t} \rho_{\mathcal{X}_{v}^{t}}$;\\
       		$\mathcal{Q}^{\prime} \leftarrow \mathcal{Q}^{\prime} \cup \{  q^{\prime}\}$;\\
       }
    	
    }
   }
   \If{$\mathcal{Q}^{\prime} \neq \emptyset$}
   {
   $q \leftarrow \argmax_ {q^{\prime}\in \mathcal{Q}^{\prime}} q^{\prime} $;\\
   $\mathcal{Q}_{\mathcal{R}^{t}}\leftarrow \mathcal{Q}_{\mathcal{R}^{t}}  \cup \{ (r^{t}_{i},q)\}$;\\ 	
   }
   
    }

        \textbf{return} $\mathcal{Q}_{\mathcal{R}^{t}}$.
     
    \caption{The Fare Estimation Algorithm}
    \label{alg:farequote}
\end{algorithm}

\subsubsection{Pickup Assignment}
Let $n_{v^{t}}$ denote the number of seats available in vehicle $v$ at time $t$. Each vehicle can only service at most $N_{v}$ passengers, where $N$ is the seat capacity of the vehicle. That is,  $ 0 \leq n_{v^{t}} \leq N_{v}$. When there are multiple requests that decrease the cost per unit demand of a coalition, the coalition selects the one that produces the lowest cost per unit demand. If there is a tie, the mechanism breaks it by choosing the one with the highest demand at random. The pickup assignment procedure is shown in Algorithm~\ref{alg:allocation}. The mechanism selects the $n_{t}$ requests that produces the lowest cost per unit demand, where $n_{t}$ is determined as following: $n_{t} = \min \{ n_{v^{t}},  n_{\mathcal{R}^{t}}  \}$, where $n_{\mathcal{R}^{t}}$ is the number of requests submitted. The time complexity of Algorithm~\ref{alg:allocation} is $O(n^{2}\log n)$.
\begin{algorithm}[h!]
    \SetKwInOut{Input}{Input}
    \SetKwInOut{Output}{Output}
    \SetKwInOut{Initialize}{Initialize}

    \underline{function assign} $(t, \mathcal{R^{\prime}}^{t})$\;
    \Input{$t$ - Time; $\mathcal{R^{\prime}}^{t}$ - A set of requests from passengers $\mathcal{I}^{t}$  who accept the fare quotes at time $t$. }
    \Output{$\Pi^{t}$-The set of assignment.}
 $\Pi^{t} \leftarrow \emptyset$; \\
 $A \leftarrow \emptyset$;\\
    \While { $v \in \mathcal{V}^{t}$}
    {
    \While{$r \in \mathcal{R^{\prime}}^{t} $} 
    {
    	\tcc{$t^{\prime}$ is the estimated pickup time needed to service passenger $i^{t}$ using  vehicle $v$ }
    	Compute $t^{\prime}$;\\
    	\If{$t^{\prime} \leq \bar{t}_{i}$}
    {
    Compute $\rho_{\mathcal{X}_{v}^{t-1} \cup \{ r\}}$ ;\\
       \If{$ \rho_{\mathcal{X}_{v}^{t-1} \cup \{ r\}} < \rho_{\mathcal{X}_{v}^{t-1}}$}
       {
    	$c^{\prime} \leftarrow \rho_{\mathcal{X}_{v}^{t}}$;\\
    	$\mathcal{A} \leftarrow \mathcal{A} \cup \{(v, c^{\prime})\}$;\\
    	}
    	}
    }

    }
    $\mathcal{V} \leftarrow \mathcal{V}^{t}$;\\
    $\mathcal{I} \leftarrow \{i \mid r_{i} \in \mathcal{R}^{t}\}$;\\
    \While{$\mathcal{A} \neq \emptyset \; {\bf  and } \; \mathcal{V} \neq \emptyset \; {\bf  and } \; \mathcal{I} \neq \emptyset$ }
    {
    \tcc{sort  in ascending order of $c^{\prime}$}
    	$\mathcal{A} \leftarrow quicksort(\mathcal{A})$;\\
    	 \tcc{ties are broken by selecting the one with the highest unit demand $\ell$}
    	$(v,c^{\prime}) \leftarrow \argmin_{(v,c^{\prime}) \in \mathcal{A}} c^{\prime}$;\\
    	$\Pi^{t} \leftarrow \Pi^{t} \cup \{(\hat{v},\hat{r})\mid \hat{v} = v, c_{\hat{r}} = c^{\prime} \}$;\\
    	$\mathcal{A} \leftarrow \mathcal{A} \setminus \{(v^{\ast}, c^{\ast}) \mid  v^{\ast} = v, (v^{\ast}, c^{\ast})\in \mathcal{A} \} $;\\
    	\If{$n_{v} < 1$}
    	{
    	$\mathcal{V} \leftarrow \mathcal{V} \setminus \{v\}$;\\
    	}
    	
    	$\mathcal{I} \leftarrow \mathcal{I} \setminus \{ i \mid c_{r_{i}} = c^{\prime}, r_{i} \in \mathcal{R^{\prime}}^{t}\}$;\\	
    }
 
      {\bf return} $\Pi^{t}$.
     
    \caption{The Pickup Assignment Algorithm}
    \label{alg:allocation}
\end{algorithm}

\subsubsection{Payment Calculation}
When a passenger accepts a fare quote and is not assigned with a vehicle,  her request will be added to time $t+1$ if the threshold $\bar{t}$ satisfies. In this process, the mechanism assumes that all the passengers accept the fare estimate. This is because if a passenger rejects the quote, the mechanism simply ignores the request and assumes that the passenger never submits it. We assume that the system can calculate the marginal cost and the optimal routes as quickly as necessary, although it might be time-consuming in real-world application due to limited computational power and the complexity of the traffic dynamics. However, it can be computed with meta heuristics~\cite{hansen2007anytime}.  At time $t$, the cost per unit demand for all requests assigned to vehicle $v$ under the assignment of $\pi_{v} \in \Pi^{t}$ is determined as following:
\begin{equation}
\rho_{\mathcal{X}^{t}_{v}} = \frac{\sum_{t^{\prime} \in [0, t]}\sum_{r \in \{r\mid (v,r)\in \Pi^{t^{\prime}}\}} \delta_{r}}{\sum_{t^{\prime} \in [0, t]}\sum_{r\in \{r\mid (v,r) \in \Pi^{t^{\prime}}\}} \ell_{r}} .
\label{eq:costperunit}
\end{equation} 

Therefore, the final payment of passenger $i$ at time $t$ is calculated as following:
\begin{equation}
p_{i}^{t} = \ell_{i}\rho_{\mathcal{X}^{t}_{v}} , 
\label{eq:payment}
\end{equation}
where $p_{i}^{t}$ can be calculated in $O(n^{2}T\log n)$ time.
\subsubsection{Ex-post Incentive Compatibility}
We show that the IORS mechanism satisfies  ex-post incentive compatibility.
\begin{lemma}
\label{le:delay}
A passenger can not decrease her cost by delaying the submission of the request, provided that all other passengers report their demand truthfully and do not change their decisions on fare quotes. That is, for all $  \tau_{1}, \tau_{2}, t \in \mathcal{T}$ and submissions $\mathbb{R}$ and $\mathbb{R}^{\prime}$, where $0\leq \tau_{1} < \tau_{2} \leq t$,  $\mathbb{R} = \{\mathcal{R}^{0}, ..\mathcal{R}^{\tau_{1}}, \mathcal{R}^{\tau_{2}}, ..., \mathcal{R}^{t}\}$, and
\begin{equation}
\mathbb{R}^{\prime}(t)= \left\{
  \begin{array}{lr}
    \mathcal{R}^{\tau_{1}} \setminus \{ r^{\tau_{1}}_{i}\} & : t = \tau_{1}\\
    \mathcal{R}^{\tau_{2}} \cup \{ r^{\tau_{1}}_{i}\} & : t = \tau_{2}\\
     \mathbb{R} (t)& : \textit{otherwise ,}\\
  \end{array}
\right.
\end{equation}
We have:
\begin{equation}
p^{t}_{\mathbb{R}(\tau_{1})} \leq p^{t}_{\mathbb{R}^{\prime} (\tau_{2})}
\label{eq:incentivecompatibility}
\end{equation}

\end{lemma}
\begin{proof}
Depending on whether $r_{i}$ is serviced or not, we distinguish two cases:
\begin{itemize}
\item The request is not serviced: if the passenger delays her request from $\tau_{1}$ to $\tau_{2}$, then her  latest departure time $\hat{\bar{\tau_{1}}} = \bar{\tau_{1}}-1 <\bar{\tau_{1}} $. If the pickup time $ t = \bar{\tau_{1}}$, then she will not be serviced at time $\tau_{2}$, which is obviously less favorable than being serviced. Another situation is that the addition of the request at time $\tau_{2}$ does not decrease the cost per unit demand of the coalitions at time $\tau_{2}$, or the new cost per unit demand is less than the threshold determined by Algorithm~\ref{alg:allocation}.
\item The request is serviced: If the passenger delays her request from $\tau_{1}$ to $\tau_{2}$.  Assuming that, 
\begin{equation}
p^{t}_{\mathbb{R}(\tau_{1})} > p^{t}_{\mathbb{R}^{\prime} (\tau_{2})}
\label{eq:contradict}
\end{equation}
We prove the theorem by contradiction.  If $0 \leq \tau < \tau_{1}$, we have $\mathbb{R}^{\prime}(\tau) = \mathbb{R}(\tau)$.  By equation~\ref{eq:demandind} and~\ref{eq:independentcost},  the operational cost and the total unit demand are independent of the request submission time.  That is, $C^{\prime} = C$, $\sum \ell^{\prime} = \sum \ell$.  By equation~\ref{eq:deltaequal}, \ref{eq:costperunit} and \ref{eq:payment},  we have $p^{t}_{\mathbb{R}(\tau)} = p^{t}_{\mathbb{R}^{\prime} (\tau)}$. Thus, inequation~\ref{eq:contradict} does not hold. This is also true if $\tau_{2} < \tau \leq t$. If $\tau_{1} <\tau \leq \tau_{2}$, since $\mathbb{R}^{\prime} (\tau) =  \mathcal{R}^{\tau_{1}} \setminus \{ r^{\tau_{1}}_{i}\}$, we have the cost per unit demand $\rho_{\mathbb{R}(\tau_{1})} \leq \rho_{\mathcal{R}^{\tau_{1}} \setminus \{ r^{\tau_{1}}_{i}\}}$, and the total demand  $\ell_{\mathbb{R}(\tau_{1})} < \ell_{\mathcal{R}^{\tau_{1}} \setminus \{ r^{\tau_{1}}_{i}\}}$. By multiplying the left and right sides of the two inequations, we get $\rho_{\mathbb{R}(\tau_{1})}\ell_{\mathbb{R}(\tau_{1})} \leq  \rho_{\mathcal{R}^{\tau_{1}} \setminus \{ r^{\tau_{1}}_{i}\}}  \ell_{\mathcal{R}^{\tau_{1}} \setminus \{ r^{\tau_{1}}_{i}\}}$. That is, $p^{t}_{\mathbb{R}(\tau)} \leq p^{t}_{\mathbb{R}^{\prime} (\tau)}$. Hence, inequation~\ref{eq:contradict} does not hold when $\tau_{2} < \tau \leq t$.  Therefore,  the assumption is invalid and inequation~\ref{eq:incentivecompatibility} holds.
\end{itemize}
By incorporating the above cases, we prove the lemma.
\end{proof}

\begin{lemma}
\label{le:waiting}
The passenger can not decrease her cost by misreporting its latest departure time, provided that all other passengers report their demand truthfully and do not change their decisions on fare quotes. That is, 
\begin{equation}
p^{\prime}_{r} \leq  p_{r}, \quad (\forall \hat{\bar{t}} \neq \bar{t}).
\label{eq:misreport}
\end{equation}
\end{lemma}
\begin{proof}[Proof sketch]
If passenger claims an earlier latest departure time (i.e., $\hat{\bar{t}} < \bar{t}$), according to Algorithm~\ref{alg:farequote} and~\ref{alg:allocation}, the search space of the vehicles may be reduced and the request might be rejected. If $\hat{\bar{t}} > \bar{t}$, the search space will be increased. However, a passenger will reject the assignment if the pickup time exceeds the $\bar{t}$ according to the assumption made in section~\ref{sec:prelim}. By equation~\ref{eq:demandind},  \ref{eq:independentcost}, \ref{eq:deltaequal}, \ref{eq:costperunit} and \ref{eq:payment}, the fare $\hat{\bar{p}}$ does not increase in either scenario.
\end{proof}

\begin{theorem}
The IORS mechanism is ex-post incentive compatible provided that all other passengers report their demand truthfully and do not change their decisions on fare quotes.
\end{theorem}
\begin{proof}[Proof sketch]
A passenger $i$ can not lie about her origin $o_{ui}$ and destination $d_{i}$. She is unable to claim an earlier arrival $t_{i}$. According to Lemma~\ref{le:delay}, she will not benefit from delaying the request submission provided that all other passengers report their demand information truthfully and do not change their decisions on whether to accept the quotes or not. The passenger will not gain from misreporting the latest departure time according to lemma~\ref{le:waiting}. 
\end{proof}

\subsubsection{Discussion}
Note that the IORS mechanism does not require passengers to specify the deadlines for the latest delivery to their destinations. This is because passengers are likely to misreport the deadlines to rule out potential ridesharing assignments (e.g., by claiming an earlier deadline). However, mechanism designers may set constraints (e.g., the longest time, the maximum number of passengers, and the maximum rate) on a shared ride if necessary.

The IORS mechanism also satisfies other properties such as individual rationality and budget balance. For example, it is individual rational because passengers' final payments never exceed their quotes. The budget balance property is met for the reason that the total cost is split by the passengers who are provided with the ride services. Due to space limitations, we omit the poofs for these properties.

\section{Benchmark}
\label{sec:benchmark}
For evaluation purposes, we compute the optimal assignment as a benchmark to evaluate the efficiency of the IORS mechanism. The goal of the optimal assignment is to minimize the overall cost per unit demand (equivalent to maximizing $W$) under the constraints in the AMoD system. This is a minimum maximal matching problem, which is NP-hard~\cite{hopcroft1973n}, and cannot be solved in polynomial time. We use  a linear programming solver \emph{LpSolve} solver~\cite{berkelaar2004lpsolve} for optimization in the experiment.

Auction-based mechanisms have been proven to be efficient in some of the existing ridesharing systems such as carpooling and shuttles~\cite{kleiner2011mechanism,cheng2014mechanisms,coltin2013towards,kamar2009collaboration}. Some of them even have very close performance compared with the optimal solution~\cite{cheng2014mechanisms}.  In our work, we compare the performance of the IORS with the state-of-the-art, offline, auction-based mechanism (bottom-up) described in~\cite{cheng2014mechanisms}. The auction-based mechanism can not be solved in polynomial time.
\section{Experimental Results}
To evaluate the performance of the IORS mechanism, we developed an AMoD  simulator to model the transportation system of a grid city with $101 \times 101$ blocks (a scenario similar to Figure~\ref{fig:abstractframework}).
\subsection{Experimental Settings}
\label{sec:expsetup}
In the experiment, we assumed that the number of AVs in the system is fixed. We set the number $N = 1000$. For each simulation, the system ran for $500$ rounds unless specified otherwise.  For each round, we generated a random number of requests $\mathcal{R}_{t} \in \mathbb{R}$. 
We initialized  $\mathbb{R}$ with a set of  $N = 500$  integers  randomly drawn from a normal distribution with mean  $\mu = 1000$ and standard deviation $100$ (see Figure~ \ref{fig:demanddistro}).  We assumed that each AV can transport up to four passengers at the same time.

We then generated the $\mathcal{R}_{t}$ requests respectively using the following method: the request time is the current round number; the waiting time is randomly drawn from the range 10 to 100;  both the origins and destinations are randomly selected within a radius of 50 blocks in the grid.  The operational cost per unit distance (block) is 1. The speeds for all vehicles are the same: 0.5 block per unit time (round).  Initially, all the AVs depot at the center of the grid city. At time $t = 0$,  the AVs become available for servicing passengers.

To be fair for evaluation and in the interest of saving time, we calculated the shortest paths between any two vertices in the city grid using the $A^{\ast}$ algorithm~\cite{hart1968formal} and saved it into a dictionary for further use in all the simulations

We ran all the simulations on a 2.9GHz quad-core machine with a 32GB RAM.
\subsection{Results and Discussion}
We performed simulations using the IORS mechanism. For comparison purposes,  we computed the optimal assignment under the same experimental settings as a benchmark.  we also conducted experiments on an AMoD system with the offline, auction-based mechanism described in section~\ref{sec:benchmark}. To counter the effect of the fluctuations caused by the randomization techniques used, we ran all the experiments 20 times and calculated the mean and the standard deviation of the metrics evaluated.

\begin{figure*}[h]
\centering
\begin{subfigure}[b]{.32\textwidth}
  \includegraphics[width=.95\linewidth]{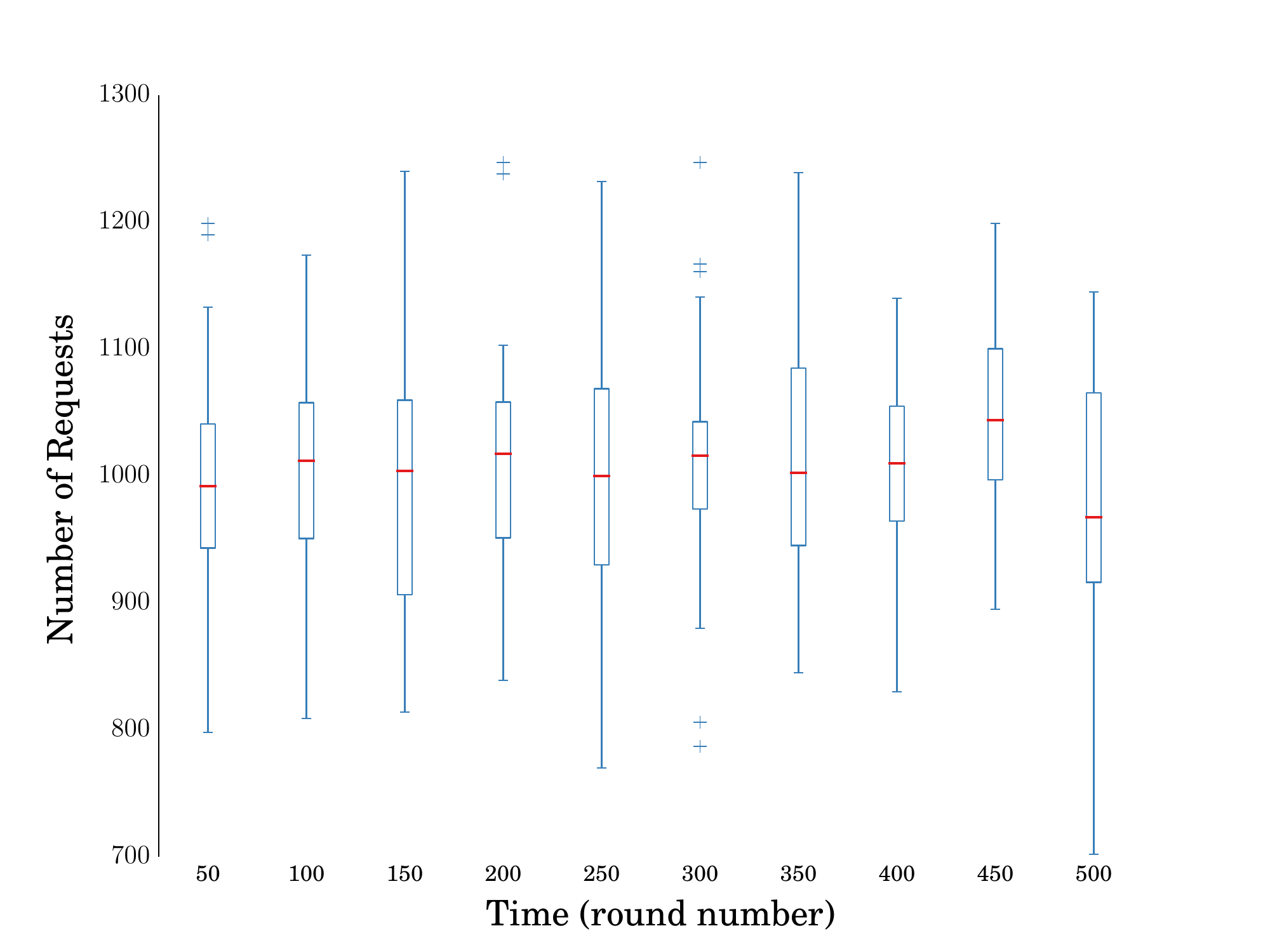}
  \caption{Demand distribution at each time.}
  \label{fig:demanddistro}
\end{subfigure}
\begin{subfigure}[b]{.32\textwidth}
  \includegraphics[width=.95\linewidth]{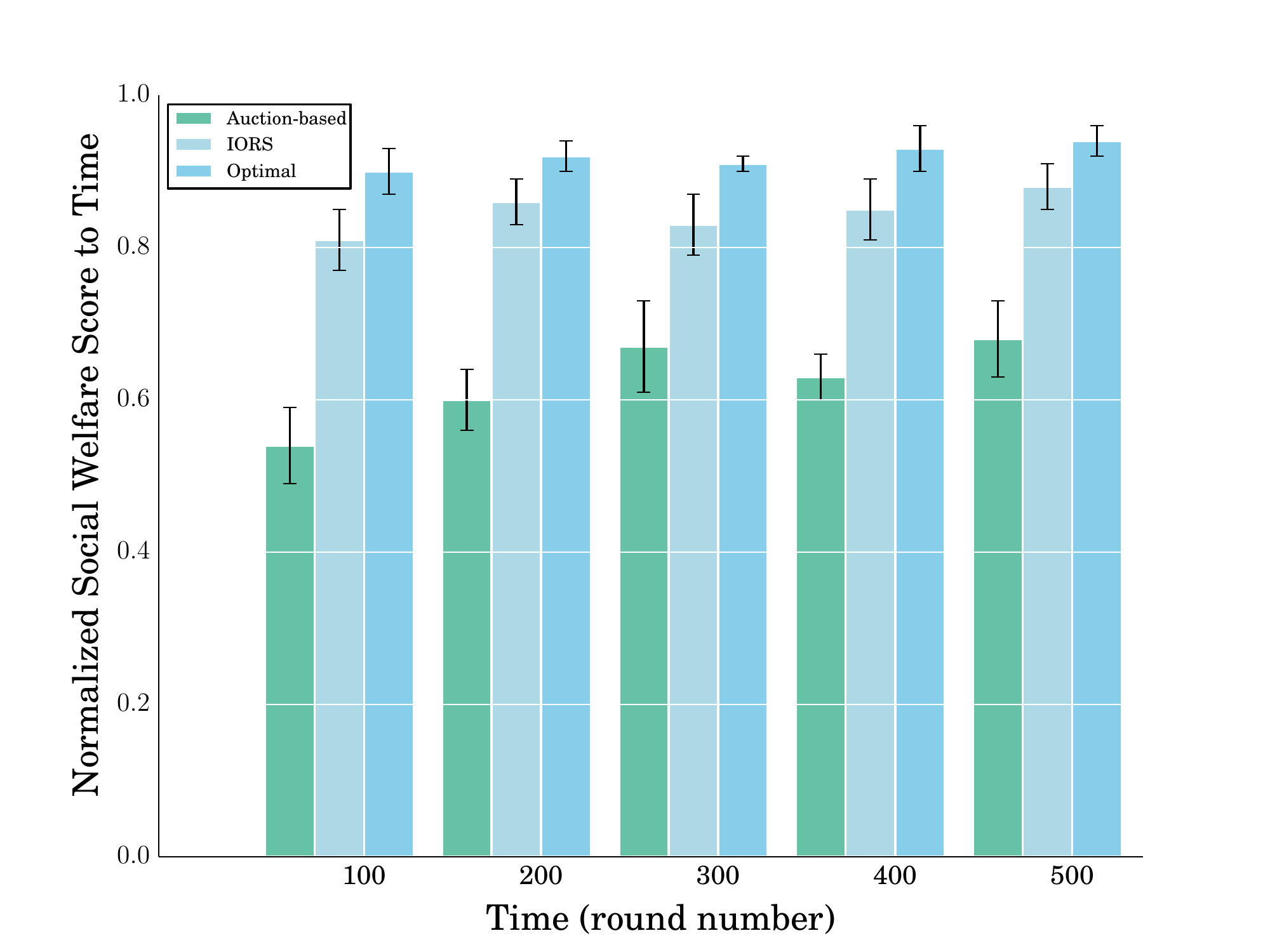}
  \caption{Social welfare scores to time.}
  \label{fig:sociawelfoverall}
\end{subfigure}%
\begin{subfigure}[b]{.32\textwidth}
\includegraphics[width=.95\linewidth]{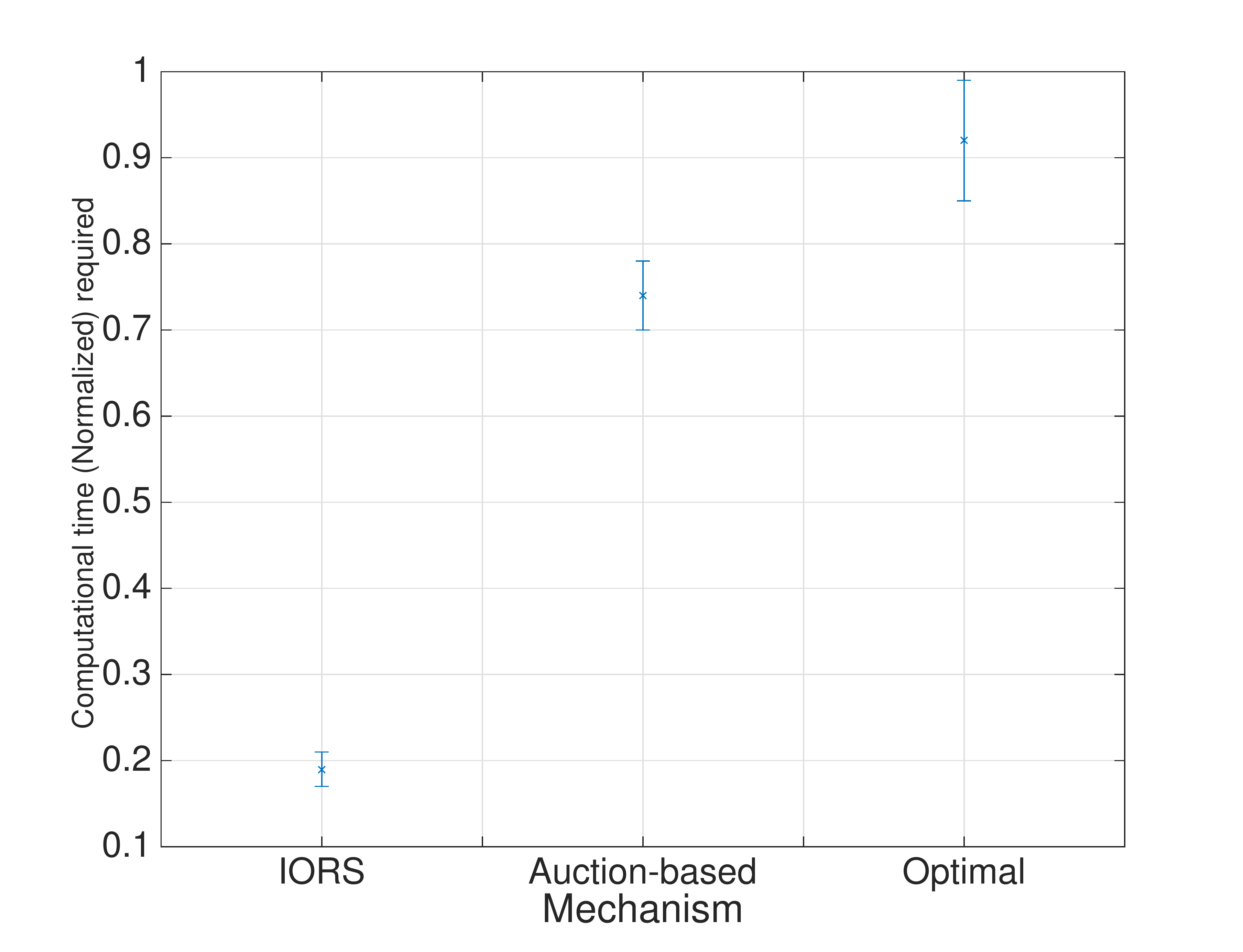}
\caption{Computational time.}
\label{fig:timecompare}
\end{subfigure}
\caption{A comparison of demand distribution, the social welfare scores and  computational time of a system with three different approaches: the IORS, an auction-based mechanism and the optimal solution.}
\label{fig:social}
\end{figure*}

We computed the social welfare scores over time for the systems using one of the following mechanisms: IORS mechanism, auction-based mechanism and the optimal assignment solution. The result (see Figure~\ref{fig:sociawelfoverall}) clearly shows that the IORS mechanism performs significantly better (with a $95\%$ confidence interval) than the auction-based mechanism, with an increase of $22.73\%$. Although it is a little inferior to the optimal solution, it performs fairly well (with a score equals to $93.62\%$ of the optimal solution) with 79.35$\%$ less computational time on average (see Figure~\ref{fig:timecompare} ) and no future knowledge of demand required. 

\begin{figure*}[h]
\centering
\begin{subfigure}[b]{.32\textwidth}
  \includegraphics[width=.95\linewidth]{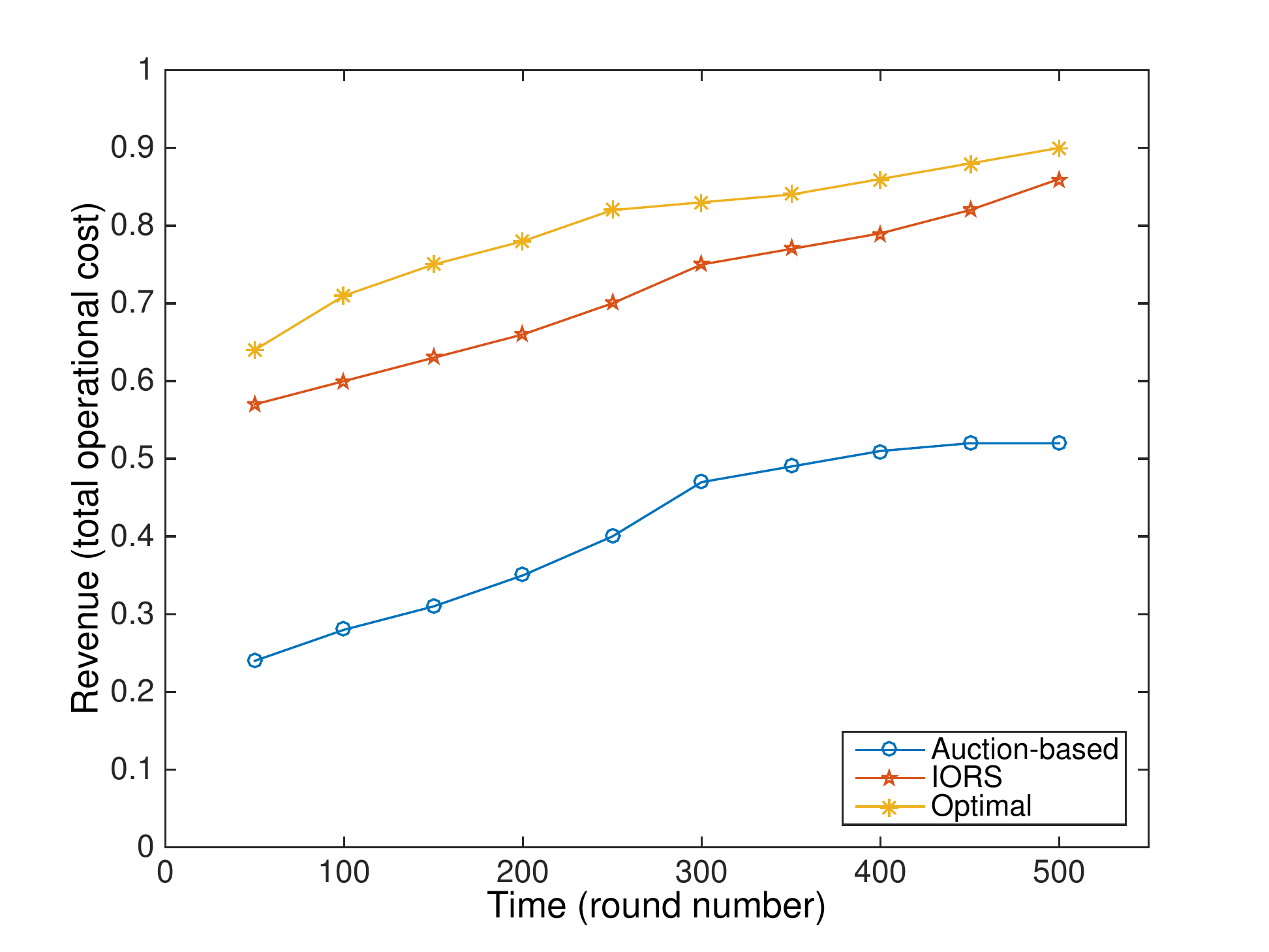}
  \caption{Revenue (total operational cost).}
  \label{fig:tripsucc}
\end{subfigure}%
~
\begin{subfigure}[b]{.32\textwidth}
  \includegraphics[width=.95\linewidth]{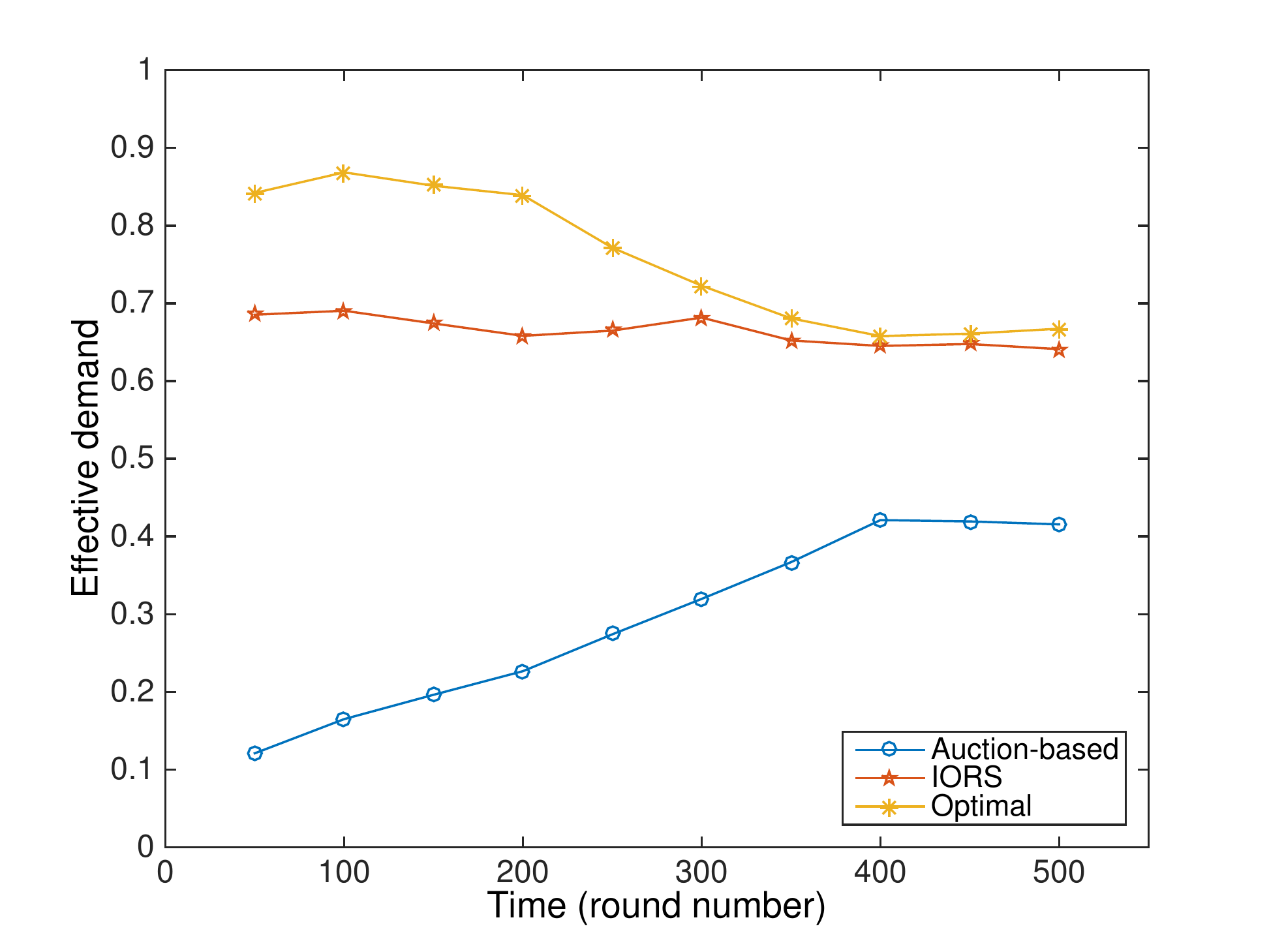}
  \caption{Total effective unit demand.}
  \label{fig:avevdt}
\end{subfigure}
\begin{subfigure}[b]{.32\textwidth}
  \includegraphics[width=.95\linewidth]{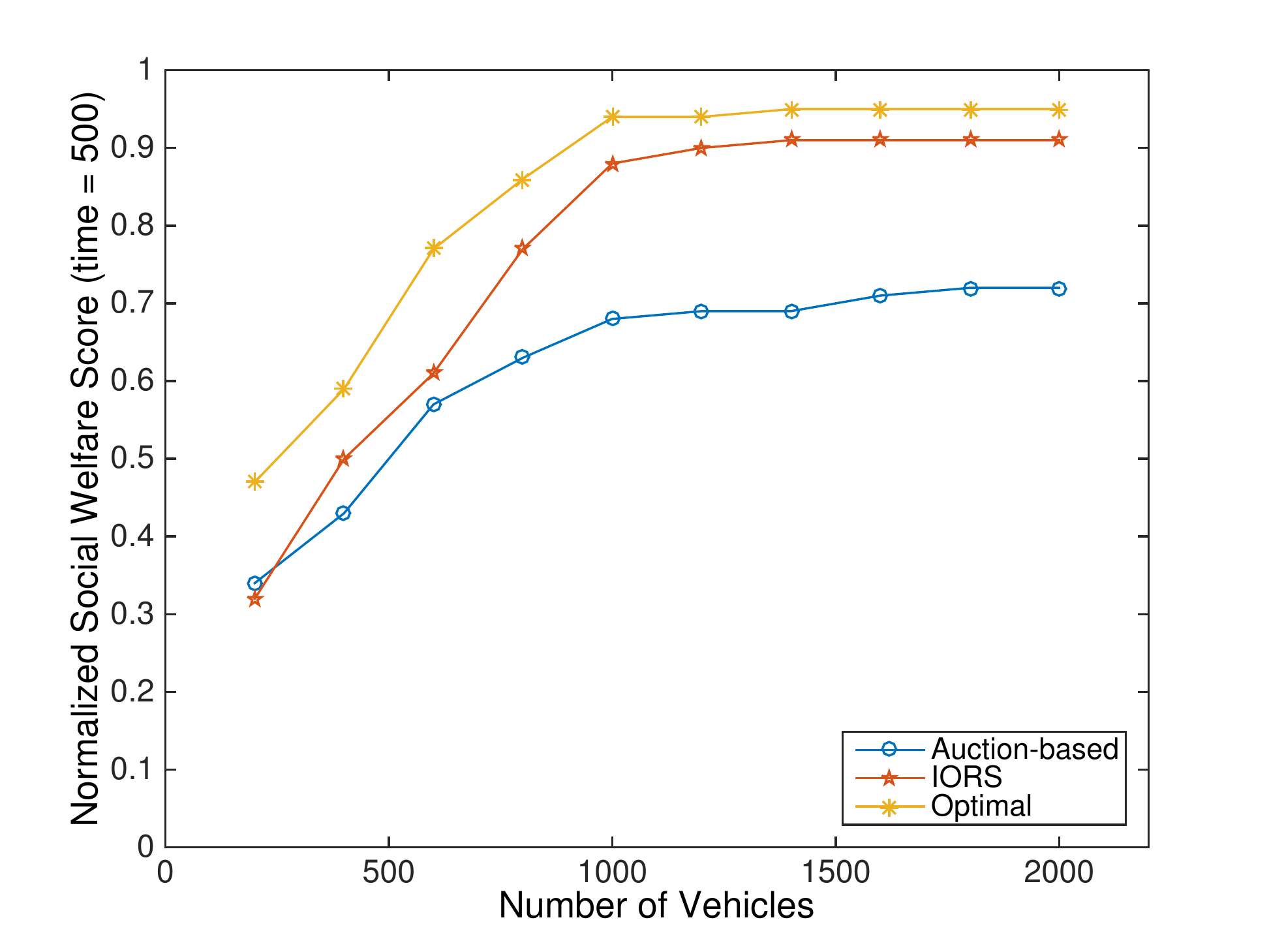}
  \caption{Varying the numbers of vehicles.}
  \label{fig:vehiclestime}
\end{subfigure}%
\caption{A comparison of the performance of a system with three approaches: the IORS, an auction-based mechanism, and the optimal solution.}
\label{fig:varingpara}
\end{figure*}
The IORS mechanism adds a request only if this addition decreases the cost per unit demand of a group. However, it might suffer from local minima and produce suboptimal solutions due to a myopic view of the demand.  The auction-based mechanism, on the other hand,  processes the aggregated requests at  once. It removes the requests with the lowest ranks.  Although the mechanism might make  better plans than the IORS mechanism at processing time because they have a better knowledge of demand distributions, it performs worse than the IORS at all the other time.

The revenue of IORS system is slightly lower than the optimal system, and much higher than the auction-based system (see Figure~\ref{fig:tripsucc}). The effective demand of the system with the IORS mechanism fluctuates around $0.7$, while the demand for optimal solution first increases and then drops down to a level very close to that of the IORS system. This is due to the increased demand from passengers.  As shown in Figure~\ref{fig:avevdt}, the demand for auction-based system keeps increasing and then reaches a plateau. For each time measured,  the scores of the auction-based system are the lowest. 

When the demand is high, obviously, it is effective to increase the supply (i.e., number of vehicles) at first. However, once the number of the AV fleet reaches some point, it will not help to improve the social welfare (i.e.,$\mathit{W}$) (see Figure~\ref{fig:vehiclestime}).

In summary, the IORS mechanism outperforms the offline, auction-based mechanism overwhelmingly in promoting ridesharing in AMoD systems. Although it is still inferior to the optimal solution, it can achieve a very close performance with substantially less computational time needed and no future knowledge of demand required.
\section{Conclusions}
To promote ridesharing in AMoD systems, we introduce a posted-price, integrated online mechanism, namely IORS. We show that IORS is ex-post incentive compatible.  Simulation results demonstrate its superiority compared with the optimal assignment solution and the offline, auction-based mechanism. Although IORS is tailored for AMoD systems, it is applicable to traditional demand responsive transport systems such as taxis and shuttles,  provided that the dispatchers have full control over the vehicles. Besides, IORS can be applied to distributed scenarios by dividing a city into multiple zones where each zone has a control center individually processing the requests.

Future directions include coalition structure generation for optimal groups of shared riders, mechanism design to address ethics and privacy problems in ridesharing. Another direction is to develop more complex and realistic simulation platforms as benchmarks for future evaluation.
\section*{Acknowledgments}
This work was supported in part by the National Science Foundation under grant CCF-1526593. The authors wish to thank anonymous reviewers for their helpful comments and suggestions.


\bibliographystyle{named}
\bibliography{ijcai16}

\end{document}